\documentclass[10pt,twocolumn,letterpaper]{article}
\usepackage[T1]{fontenc}
\usepackage{newtxtext,newtxmath}
\usepackage[margin=0.75in]{geometry}
\usepackage[hyphens]{url}
\usepackage{graphicx}
\usepackage[numbers,sort&compress]{natbib}
\usepackage{caption}
\usepackage{booktabs}
\usepackage{algorithm}
\usepackage{algpseudocode}
\usepackage{amsmath}
\usepackage{float}
\usepackage{placeins}
\usepackage{multirow,tabularx,array,xcolor}
\usepackage{microtype}
\usepackage{hyperref}
\title{Revealing and Bridging the Physical Executability Gap in Vision-Language Navigation}

\author{Gang He\textsuperscript{1} \quad Zhile Cheng\textsuperscript{1} \quad
Kepeng Xu\textsuperscript{1}\thanks{Corresponding author: \href{mailto:kepengxu11@gmail.com}{kepengxu11@gmail.com}.} \\
Zhenyang Liu\textsuperscript{1} \quad Yu-Wing Tai\textsuperscript{2} \\[0.5em]
\small \textsuperscript{1}Xidian University \quad \textsuperscript{2}Dartmouth College}
\date{}

\begin{document}
\raggedbottom

\maketitle

\begin{abstract}
Continuous vision-and-language navigation (VLN) is typically evaluated by goal-reaching success, assuming it reflects an agent's ability to execute language-grounded behavior. We show that this assumption can fail: permissive contact models in current simulators may mask physically infeasible actions, creating a physical executability gap between semantic competence and embodied realization. To bridge this gap, we introduce Visual-Semantic Traversability Alignment (VeSTA), a framework that aligns VLN policies with physically executable futures. We define visual-semantic traversability as the ability to predict instruction-conditioned trajectories that are semantically valid, geometrically feasible, and physically robust. VeSTA addresses this misalignment through two complementary mechanisms: improving executable support in trajectory generation and risk-aware selection among multimodal futures. Specifically, Trajectory Relative Group Policy Optimization uses strict-contact feedback to reshape the candidate distribution toward executable regions, while the Traversability Risk Critic learns a risk-adjusted scalar utility for candidate-wise future selection. Experiments on R2R and RxR reveal substantial dependence of existing VLN agents on simulator contact semantics. Under strict-contact evaluation, VeSTA substantially reduces collision-induced failures while preserving or improving goal-reaching performance, and controlled real-world robot experiments provide evidence that the learned preference transfers beyond simulation.
\end{abstract}

\section{Introduction}

Vision-and-language navigation (VLN) maps natural-language instructions and visual observations to goal-directed behavior. Whereas R2R formulates navigation over discrete graphs \citep{anderson2018vision}, continuous VLN must further realize semantic decisions through low-level motion in 3D environments \citep{krantz2020beyond,savva2019habitat}. Navigation success therefore depends on two related but distinct capabilities: identifying a semantically appropriate goal and converting that intention into motion that can be executed under local geometric and physical constraints.

\begin{figure}[!t]
\centering
\includegraphics[width=\columnwidth,height=0.565\columnwidth]{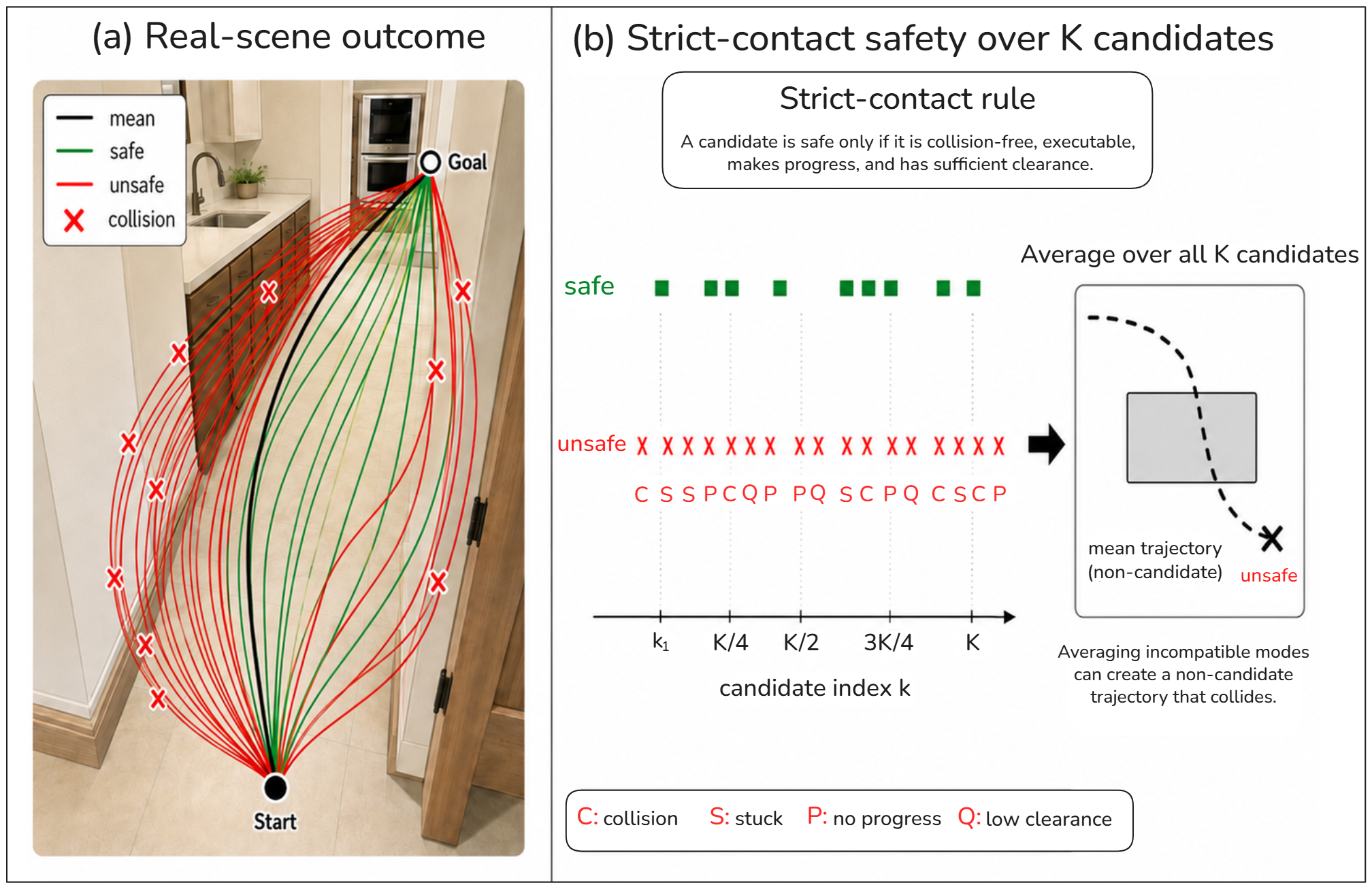}
\caption{Conceptual illustration: averaging incompatible modes can yield a non-candidate trajectory that collides under strict contact; not all baselines use literal mean aggregation.}
\label{fig:mean_candidate_failure}
\end{figure}

\begin{figure*}[!t]
\centering
\includegraphics[width=\textwidth,height=0.443\textwidth]{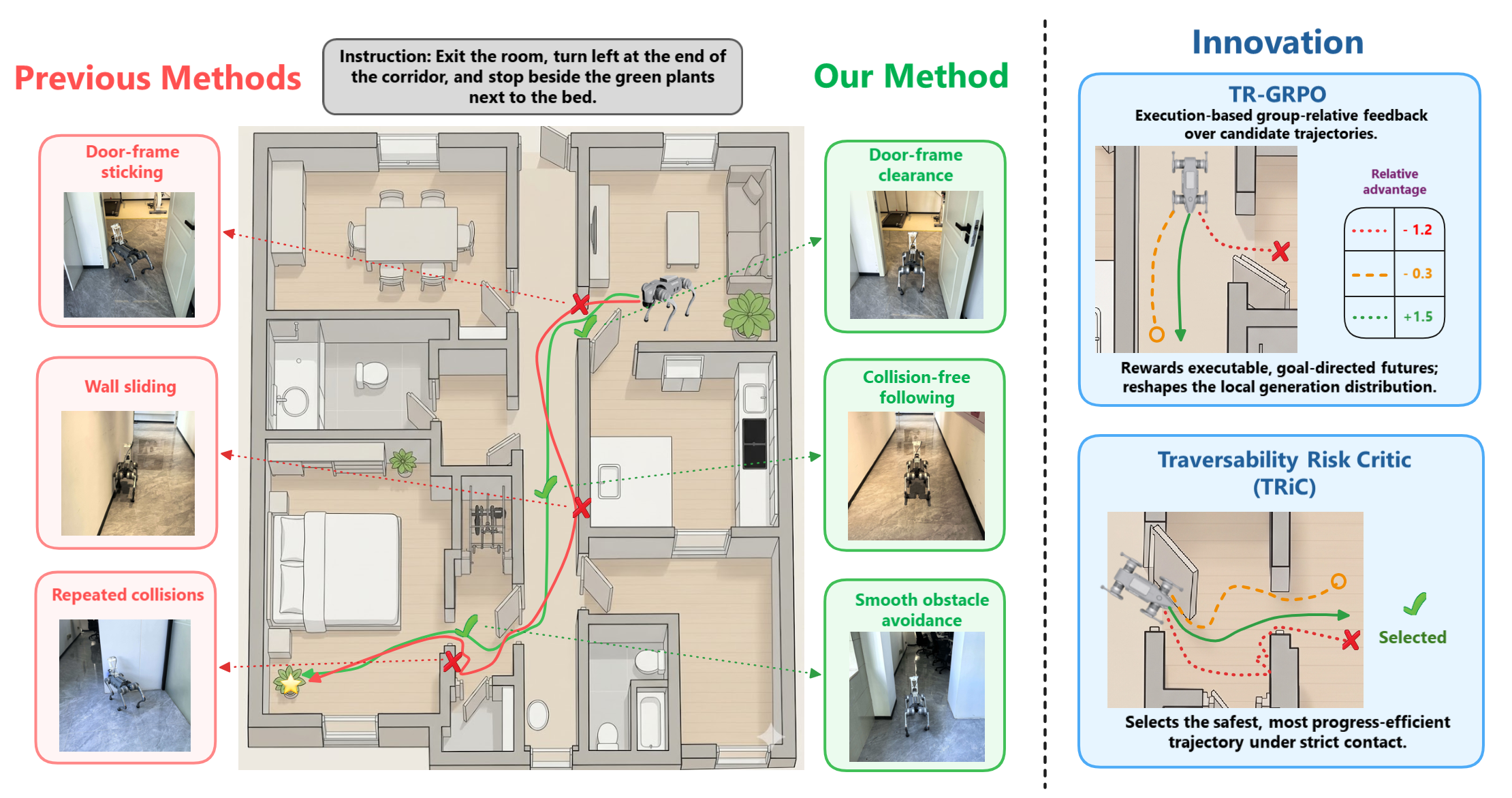}
\caption{Overview of VeSTA. Trajectory Relative Group Policy Optimization uses strict-contact feedback to reshape candidate support toward executable regions. The Traversability Risk Critic learns a scalar utility from progress, contact, and immobilization preferences for candidate-wise receding-horizon selection.}
\label{fig:method}
\end{figure*}

Permissive contact resolution can obscure this distinction: an obstacle-directed command may be projected into tangential displacement, yielding apparent progress without a realizable local motion, a simulator behavior previously shown to distort sim-to-real navigation comparisons \citep{kadian2020sim2real}. We call the discrepancy between semantic intent and executable behavior the \emph{physical executability gap}. We define \emph{visual-semantic traversability} as predicting instruction-conditioned trajectories that are semantically valid, geometrically feasible, and robust to persistent contact or immobilization; it is not a claim of general safety.

We diagnose this gap by changing only contact semantics. Strict contact blocks obstacle-directed motion rather than repairing it through tangential progress. With policies and episode specifications fixed, the resulting performance change measures system-level sensitivity, not the pointwise causal effect of a collision. This intervention separates two questions that goal-reaching metrics conflate: whether the agent chooses an instruction-consistent direction and whether its local motion can actually realize that choice. We formulate local execution alignment through two complementary mechanisms: \emph{trajectory distribution bias}, which governs executable support in the candidate distribution, and \emph{multimodal future selection uncertainty}, which governs utility-based choice among incompatible candidates. The former determines which futures are proposed; the latter determines which future is executed.

VeSTA augments a dual-system VLN agent \citep{dualvln2025} while keeping its high-level planner fixed. TR-GRPO uses strict-contact feedback and a local action-marginal surrogate to shape executable candidate support; TRiC learns a scalar utility from multi-factor trajectory preferences to select an individual future. We evaluate this design through contact intervention, R2R/RxR benchmarks, ablations, and controlled robot trials.

Our contributions are fourfold. First, we expose contact-semantics sensitivity as a diagnostic of the physical executability gap. Second, we formalize visual-semantic traversability through complementary alignment of candidate generation and multimodal selection. Third, VeSTA combines TR-GRPO for executable-support formation with TRiC for candidate-level utility. Finally, strict-contact benchmarks, component ablations, and robot trials show substantially lower collision burden without reducing reported goal-reaching performance.

\begin{figure*}[!t]
\centering
\includegraphics[width=0.98\textwidth]{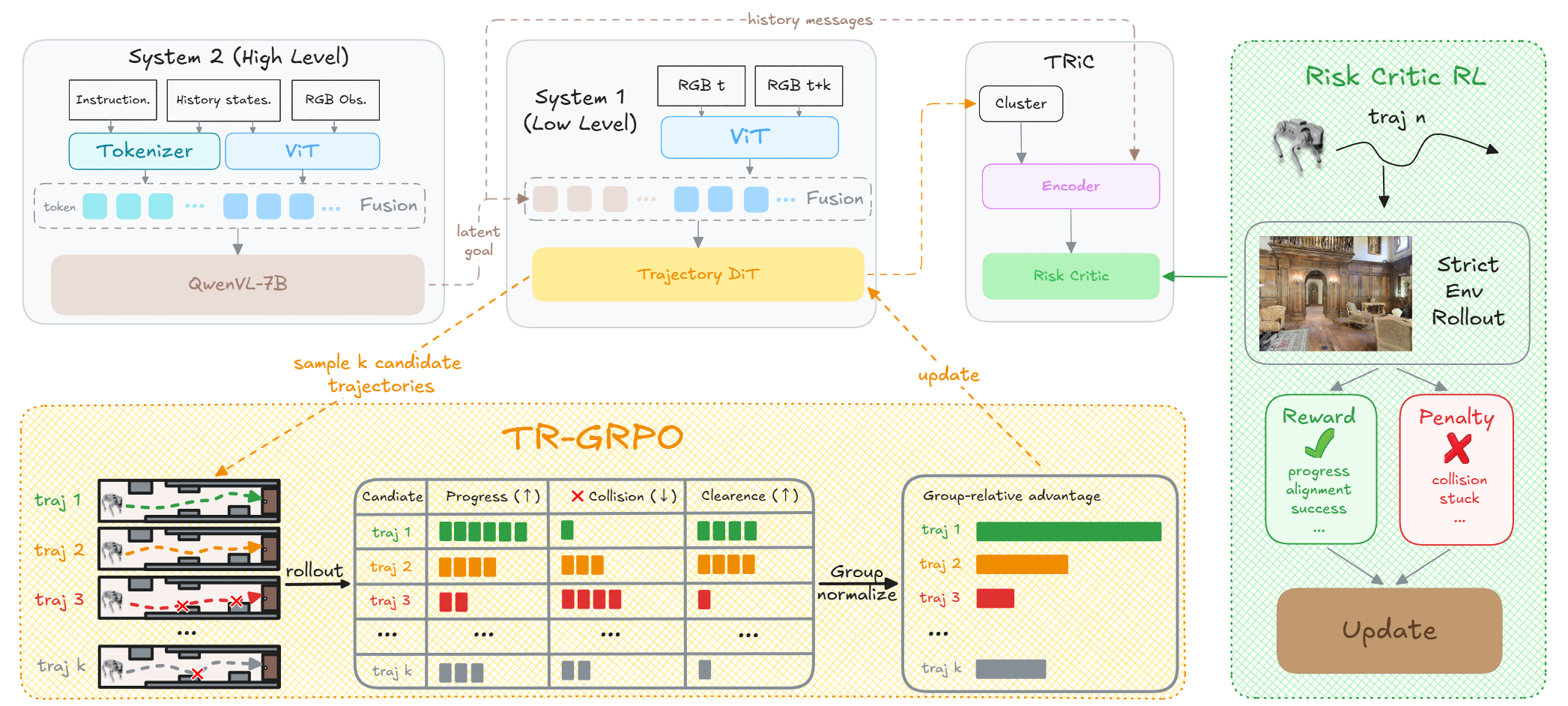}
\caption{Contact-semantics intervention. Existing agents degrade when obstacle-directed motion is blocked; VeSTA reduces collision burden through distribution alignment and risk-aware selection.}
\label{fig:motivation}
\end{figure*}

\section{Related Work}

\paragraph{Continuous VLN.} Continuous VLN extends graph-based instruction following to embodied control in 3D environments \citep{anderson2018vision,krantz2020beyond,savva2019habitat}. Language-conditioned waypoint, topology, map, and world-model representations include Topological Planning with Transformers, Waypoint Models, discrete-to-continuous transfer, CM2, BEVBert, GridMM, DREAMWALKER, and NavMorph \citep{pashevich2021topological,krantz2021waypoint,hong2022bridging,georgakis2022crossmodal,an2023bevbert,wang2023gridmm,wang2023dreamwalker,ghaus2025navmorph}. Recent video/VLM systems include NaVid, NaVILA, Uni-NaVid, StreamVLN, JanusVLN, AwareVLN, and EvoNav \citep{zhang2024navid,cheng2024navila,zhang2024uninavid,streamvln2025,janusvln2025,guo2026awarevln,dai2026evonav}. Complementary work scales synthetic VLN training or introduces explicit map- and LLM-guided planning, including ScaleVLN, MapGPT, and NavGPT-2 \citep{wang2023scalevln,chen2024mapgpt,zhou2024navgpt2}. These methods improve semantic reasoning, memory, data coverage, or local motion proposals; our complementary question is whether the selected intention remains executable when contact-prone commands are no longer repaired.

\paragraph{Multimodal trajectory futures.} Candidate-based forecasting preserves distinct futures instead of collapsing them into a single regression target. DESIRE samples, ranks, and refines diverse trajectories, while MultiPath represents multiple modes through learned anchor trajectories \citep{lee2017desire,chai2020multipath}. Although developed for dynamic-scene forecasting rather than instruction following, they motivate VeSTA's candidate-wise treatment of incompatible local futures; TRiC differs by scoring instruction-conditioned robot trajectories using strict-contact execution feedback.

\paragraph{General visual navigation policies.} Beyond VLN, ViNT learns a reusable visual navigation backbone across robots and environments, while NoMaD uses goal-masked diffusion for navigation and exploration \citep{shah2023vint,sridhar2024nomad}. Diffusion Policy similarly demonstrates multimodal visuomotor action generation through denoising \citep{chi2023diffusion}. These systems are not substitutes for language grounding, but they motivate RGB-conditioned, multimodal local policies; VeSTA focuses on aligning such local futures with an instruction-derived subgoal and strict-contact executability.

\paragraph{Collision-aware execution.} Classical planners optimize feasible velocities or short-horizon trajectories under geometry \citep{fox1997dynamic,rosmann2013efficient}. Safety-aware or execution-aware VLN may instead add LiDAR fallback, repulsive collision estimation, language-model-assisted correction, or an obstacle-avoiding controller \citep{yue2023safevln,kim2025care,jeong2024vlncm,an2025etpnav}. Many specialized systems therefore use extra range sensing or an external correction stage, although this is not universal. VeSTA augments a general-purpose VLN policy with RGB-only inference and no added LiDAR/depth or external local planner. This comparison distinguishes sensing assumptions and design scope rather than claiming superiority over methods targeting broader notions of safety. Recent physically grounded VLN evaluation further documents collisions, falls, observation shifts, and embodiment-dependent locomotion constraints that are absent from idealized point-agent evaluation \citep{wang2025embodiedgap}.

\paragraph{Hierarchical RL.} DualVLN separates semantic planning from trajectory generation \citep{dualvln2025}, allowing us to hold planning fixed and isolate local execution. This modularity is important experimentally: changing only S1 avoids attributing improved contact behavior to a stronger language planner. TR-GRPO builds on flow matching \citep{lipman2023flow}, flow-policy fine-tuning \citep{lyu2025fpo}, and group-relative optimization \citep{shao2024deepseekmath,li2026navgrpo}. NavGRPO uses group-relative learning to promote robust and diverse instruction-following trajectories, but evaluates graph-based VLN; our objective instead uses strict-contact feedback for continuous local execution. Candidates share a visual-language state, and a collapsed action marginal provides a tractable local ratio that acts directly on the VLN policy rather than through an external recovery controller.

\section{Method}

\subsection{Executability Diagnostic and Setup}
Let $s_t$, $a_t$, and $\kappa$ denote state, local command, and contact semantics:
\begin{equation}
\label{eq:contact_transition}
s_{t+1}=\mathcal{T}_{\kappa}(s_t,a_t).
\end{equation}
Permissive semantics may project an obstacle-directed command tangentially, whereas strict semantics blocks it. For a fixed policy $\pi$ and navigation metric $M$, contact sensitivity is
\begin{equation}
\label{eq:contact_gap}
\Delta_{\kappa}M(\pi)=M(\pi;\kappa_{\mathrm{perm}})-M(\pi;\kappa_{\mathrm{strict}}).
\end{equation}
This indicator is a system-level diagnostic rather than a new benchmark score. A positive SR or SPL gap indicates that some measured goal-reaching performance depends on permissive realization of contact-prone commands; increased strict-contact collision burden locates the discrepancy at the intent--execution interface. Because contact changes later observations and episode length, $\Delta_{\kappa}M$ is not a pure causal estimate of an individual collision. VeSTA does not attempt to reproduce tangential projection inside the policy. Instead, it favors trajectories whose intended progress remains realizable when environment-side correction is absent.

Given observation $o$ and fixed high-level subgoal $g$, S1 defines $\pi_\theta(\tau\mid o,g)$ over short-horizon trajectories and samples
\begin{equation}
\label{eq:candidate_set}
\mathcal{C}=\{\tau^1,\ldots,\tau^K\}.
\end{equation}
The \emph{executable motion manifold} denotes, conceptually, the state-conditioned trajectory region that makes useful instruction-directed progress without persistent contact or immobilization; no smooth differential-manifold assumption is required. Executability therefore depends jointly on observation, subgoal, embodiment, local geometry, and motion representation. For strict-contact utility $V$, selection regret is
\begin{equation}
\label{eq:selection_regret}
\mathcal{R}_{\mathrm{sel}}=\max_{\tau\in\mathcal{C}}V(\tau\mid o,g)-V(\hat{\tau}\mid o,g).
\end{equation}
TR-GRPO improves the support from which candidates are drawn, while TRiC reduces $\mathcal{R}_{\mathrm{sel}}$ when mapping that set to one action. This functional decomposition assigns complementary optimization roles to generation and selection: the former determines which futures are proposed, and the latter determines which future is executed. Holding S2 fixed isolates local execution alignment from high-level language reasoning and allows VeSTA to refine an existing dual-system agent.

\subsection{TR-GRPO for Trajectory Distribution Alignment}
For candidates sampled under the same $(o,g)$, a strict-contact probe produces
$r^i=\lambda_pP(\tau^i)-\lambda_cC(\tau^i)-\lambda_uU(\tau^i)$, where $P$ aggregates goal progress, success, and clearance benefit, $C$ measures collision frequency, and $U$ aggregates clearance risk, immobilization, and unsafe-candidate penalties. The group-relative advantage
\begin{equation}
\label{eq:group_relative_advantage}
A^i=\frac{r^i-\mu_r}{\sigma_r+\epsilon_n}
\end{equation}
ranks alternative realizations while controlling for state difficulty. A candidate $\tau=(\delta x_h,\delta y_h,\delta\psi_h)_{h=1}^{H}$ is represented by the controller-facing trajectory vector $a(\tau)=\mathrm{vec}(\tau)\in\mathbf{R}^{3H}$. Because the flow sampler does not expose an exact path density, TR-GRPO defines the diagonal local policy
\begin{equation}
\label{eq:collapsed_action_marginal}
\tilde{\pi}_\theta(a\mid o,g)=\mathcal{N}\!\left(a;\mu_\theta(o,g),\mathrm{diag}(\sigma_\theta^2(o,g))\right).
\end{equation}
Here $\mu_\theta$ is the vectorized deterministic S1 trajectory, while $\sigma_\theta$ is produced by a Gaussian scale head from the same observation-conditioned policy feature. Consequently, log-probability gradients update both the S1 mean-producing modules and the exploration scale. Following clipped policy optimization \citep{schulman2017ppo},
\begin{equation}
\label{eq:proxy_grpo_objective}
\begin{array}{rcl}
\rho^i&=&\displaystyle\frac{\tilde{\pi}_\theta(a(\tau^i)\mid o,g)}{\tilde{\pi}_{\theta_{\mathrm{old}}}(a(\tau^i)\mid o,g)},\\
\mathcal{J}_{\mathrm{TR}}(\theta)&=&\displaystyle\mathrm{E}\!\left[\min\!\left(\rho^iA^i,\mathrm{clip}(\rho^i,1-\epsilon,1+\epsilon)A^i\right)\right].
\end{array}
\end{equation}
In the reported model, $H=6$, two reparameterized candidates are probed from the same restored simulator state by executing their first four steps, and $\epsilon=0.2$. Probe weights are $1$ for progress, local success, success-zone, clearance-risk, and stuck terms, $0.25$ for clearance, $3$ for collision, and $10$ for terminal success; candidates with any collision, clearance below $0.15$ m, or no progress receive an additional $100$ penalty. The term $-0.1\mathcal{J}_{\mathrm{TR}}$ is added to the standard PPO actor--critic loss. Thus complete-trajectory contact feedback determines candidate preference, while the diagonal trajectory-vector policy provides its differentiable optimization interface.

\begin{algorithm}[t]
\caption{TR-GRPO (S1 sampler training)}
\label{alg:proxygrpo}
\small
\begin{algorithmic}[1]
\Require S1 sampler $\pi_\theta$, strict-contact probe $\Phi$, group size $G$, clip $\epsilon$, weight $\eta$, optimizer
\State $\theta_{\mathrm{old}} \leftarrow \theta$
\Repeat
  \State sample observation $o$ and high-level subgoal $g \leftarrow \mathrm{S2}(o)$
  \State sample $\{\tau^i\}_{i=1}^{G} \sim \pi_{\theta_{\mathrm{old}}}(\cdot \mid o, g)$
  \State $r^i \leftarrow \Phi(\tau^i, o, g)$
  \State $A^i \leftarrow (r^i - \mathrm{mean}_i\, r^i) / (\mathrm{std}_i\, r^i + \epsilon_n)$
  \State $\rho^i \leftarrow \tilde{\pi}_\theta(a(\tau^i) \mid o, g) \,/\, \tilde{\pi}_{\theta_{\mathrm{old}}}(a(\tau^i) \mid o, g)$
  \State $\mathcal{L} \leftarrow \mathcal{L}_{\mathrm{PPO}}-\frac{\eta}{G}\sum_i \min\!\bigl(\rho^i A^i,\, \mathrm{clip}(\rho^i, 1{-}\epsilon, 1{+}\epsilon)\, A^i\bigr)$
  \State $\theta \leftarrow \mathrm{Opt.step}(\theta, \nabla_\theta \mathcal{L})$
  \State $\theta_{\mathrm{old}} \leftarrow \theta$
\Until{converged}
\end{algorithmic}
\end{algorithm}

\subsection{TRiC for Multimodal Future Selection}
Averaging incompatible modes can produce a non-candidate action (Figure~\ref{fig:mean_candidate_failure}); TRiC instead models multimodal future selection as conditional candidate ranking. Progress, contact, and immobilization in the probe utility $r^i$ define the semantic structure of the group preference. TRiC learns this preference as a unified scalar utility
\begin{equation}
\label{eq:tric_score}
S_\phi^i=f_\phi(o,g,\tau^i),
\end{equation}
with decision
\begin{equation}
\label{eq:tric_decision}
\tau^\ast=\arg\max_{\tau^i\in\mathcal{C}}S_\phi^i.
\end{equation}
Algorithm~\ref{alg:tric} trains $f_\phi$ from strict-contact group preferences, preserving the ordering induced by progress, contact, and immobilization without collapsing distinct motion modes. Candidate-wise scoring supports state-dependent trade-offs between progress and executability, and receding-horizon execution applies a prefix of $\tau^\ast$ before replanning.

\begin{algorithm}[t]
\caption{TRiC Risk Critic Training}
\label{alg:tric}
\small
\begin{algorithmic}[1]
\Require frozen S1 $\pi_\theta$, frozen S2, strict-contact probe $\Phi$, Risk Critic $f_\phi$, group size $K$, listwise temperature $\beta$, optimizer
\Repeat
  \State sample observation $o$ and subgoal $g \leftarrow \mathrm{S2}(o)$
  \State sample $\{\tau^i\}_{i=1}^{K} \sim \pi_\theta(\cdot \mid o, g)$
  \State $r^i \leftarrow \Phi(\tau^i, o, g)$
  \State $A^i \leftarrow (r^i - \mathrm{mean}_i\, r^i) / (\mathrm{std}_i\, r^i + \epsilon_n)$
  \State $S_\phi^i \leftarrow f_\phi(o, g, \tau^i)$
  \State $\pi_\phi(\tau^i \mid o, g) \leftarrow \mathrm{softmax}_i\!\bigl(S_\phi^i / \beta\bigr)$
  \State $\mathcal{L} \leftarrow -\sum_i A^i \log \pi_\phi(\tau^i \mid o, g)$
  \State $\phi \leftarrow \mathrm{Opt.step}(\phi, \nabla_\phi \mathcal{L})$
\Until{converged}
\end{algorithmic}
\end{algorithm}

\section{Experiments}

\subsection{Experimental Questions and Protocol}
The experiments address four questions: (i) does changing contact semantics reveal a physical executability gap; (ii) can VeSTA reduce collision burden while preserving instruction-conditioned navigation; (iii) how do trajectory distribution alignment and risk-aware multimodal future selection affect aggregate execution outcomes; and (iv) does the learned preference for executable motion remain applicable in physical deployment? We evaluate R2R Val-Unseen \citep{anderson2018vision} and English-instruction episodes from RxR Val-Unseen \citep{ku2020rxr}. Standard metrics are navigation error (NE), oracle success (OS), success rate (SR), and success weighted by path length (SPL) \citep{anderson2018evaluation,krantz2020beyond}; RxR additionally includes normalized Dynamic Time Warping (nDTW) \citep{ilharco2019dtw}, following the standard RxR-CE convention.

Goal-reaching metrics alone cannot distinguish semantic failure from repeated local contact. For an episode $e$ with $T_e$ executed steps and $C_e$ contact events, we define $\mathrm{CollRate}_e=C_e/T_e$, the episode-averaged burden $\mathrm{CollAvg}=\frac{1}{N}\sum_e \mathrm{CollRate}_e$, and the step-weighted burden $\mathrm{CollW}=\sum_e C_e / \sum_e T_e$. The first estimates contact propensity for a typical episode; the second assigns greater influence to long trajectories with persistent wall interaction or immobilization. Together with SR and SPL, these quantities test whether apparent progress is obtained through executable behavior.

Strict contact is the primary condition; permissive contact is an intervention baseline. Policies, episodes, stopping criteria, success radii, and budgets are fixed while only contact changes; DualVLN-family comparisons also fix S2 and the S1 architecture. The matched DualVLN comparison is therefore the primary efficacy test, while cross-architecture baseline rankings are descriptive unless every implementation detail is independently controlled. SR/SPL measure realized progress, whereas Coll. Avg. and Coll. W. measure typical and long-trajectory contact burden. Reduced collision without lost navigation is desired; no-motion is not improvement. Because tables are aggregate evaluations without repeated-seed confidence intervals, collision reductions are the primary evidence and smaller navigation-metric differences are descriptive.

\subsection{Strict-Contact Navigation and Execution Reliability}
\begin{table}[H]
\centering
\scriptsize
\setlength{\tabcolsep}{2.0pt}
\resizebox{\columnwidth}{!}{%
\begin{tabular}{lcccccc}
\toprule
Method & NE$\downarrow$ & OS$\uparrow$ & SR$\uparrow$ & SPL$\uparrow$ & Coll. Avg.$\downarrow$ & Coll. W.$\downarrow$ \\
\midrule
NaVid & 6.48 & 30.46 & 25.49 & 23.16 & 18.20 & 15.80 \\
NaVILA & 6.62 & 38.39 & 28.55 & 26.43 & 35.18 & 59.62 \\
UniNaVid & 6.16 & 35.43 & 32.26 & 30.50 & 22.46 & 21.18 \\
StreamVLN & 6.19 & 37.74 & 30.30 & 27.59 & 39.03 & 67.85 \\
JanusVLN & 6.79 & 30.40 & 24.80 & 23.39 & 54.52 & 77.64 \\
DualVLN & 4.84 & 57.64 & 51.50 & 47.61 & 13.22 & 34.00 \\
\textbf{VeSTA (Ours)} & \textbf{4.63} & \textbf{59.38} & \textbf{53.56} & \textbf{48.93} & \textbf{6.08} & \textbf{12.73} \\
\bottomrule
\end{tabular}
}
\caption{Navigation and execution results on R2R Val-Unseen under strict contact semantics. Standard navigation metrics measure goal-reaching performance, while collision metrics characterize local execution reliability without environment-side contact correction. Collision metrics are percentages.}
\label{tab:main_r2r}
\end{table}

Among the displayed strict-contact results, VeSTA has the highest SR and lowest NE and collision burden. The controlled claim rests on the matched DualVLN comparison: Coll. W. decreases from 34.00 to 12.73, while the smaller SR change from 51.50 to 53.56 is descriptive. Non-decreasing SR/SPL makes trivial no-motion unlikely, but cannot exclude subtler conservatism such as fewer steps or earlier stopping.

\begin{table}[H]
\centering
\scriptsize
\setlength{\tabcolsep}{2.0pt}
\resizebox{\columnwidth}{!}{%
\begin{tabular}{lccccccc}
\toprule
Method & NE$\downarrow$ & OS$\uparrow$ & SR$\uparrow$ & SPL$\uparrow$ & nDTW$\uparrow$ & Coll. Avg.$\downarrow$ & Coll. W.$\downarrow$ \\
\midrule
DualVLN (baseline) & 6.78 & 49.06 & 41.54 & 35.99 & 59.13 & 21.79 & 39.17 \\
\textbf{VeSTA (Ours)} & \textbf{6.48} & \textbf{50.50} & \textbf{42.41} & \textbf{36.51} & \textbf{60.45} & \textbf{9.36} & \textbf{15.53} \\
\bottomrule
\end{tabular}
}
\caption{Results on RxR Val-Unseen under strict contact semantics. The evaluation includes $N=3669$ episodes with English instructions and follows the standard RxR-CE reporting convention. Collision metrics characterize local execution reliability; OS, SR, SPL, nDTW, and collision metrics are percentages.}
\label{tab:rxr_en}
\end{table}

RxR shows the same aggregate pattern: the primary evidence is Coll. W. decreasing from 39.17 to 15.53. Smaller NE, SR, SPL, and nDTW differences are descriptive. Their non-decreasing direction is consistent with retained navigation, but trajectory-length and stopping diagnostics are needed to exclude subtler conservatism.

\vspace{-0.6em}
\subsection{Transfer to Physical Execution}
\vspace{-0.2em}
The robot study examines whether strict-contact alignment captures a transferable property of visual motion rather than a simulator-specific convention, complementing recent evidence that physical VLN deployment exposes collision and embodiment failures hidden by idealized evaluation \citep{wang2025embodiedgap}. We deploy DualVLN and VeSTA on a Unitree Go2 quadruped \citep{unitreego2} in three controlled indoor scene groups following the DualVLN protocol \citep{dualvln2025}: Hallway (Easy), Single Bedroom (Medium), and Room-to-Room (Hard). Each scene group contains 20 RGB-only instruction trials per method. Collision-Free Success Rate (CFSR) requires both goal completion and absence of physical contact, making it a direct physical analogue of the joint success--executability criterion studied in simulation.

\vspace{-0.35em}
\begin{center}
\includegraphics[width=0.72\columnwidth]{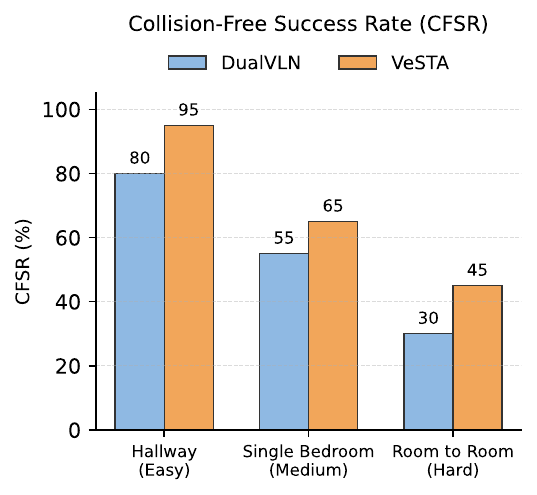}
\captionof{figure}{Collision-Free Success Rate on a Unitree Go2 in Hallway (Easy), Single Bedroom (Medium), and Room-to-Room (Hard) environments. Each method is tested on 20 instructions per scene group, and CFSR requires both goal completion and absence of physical contact.}
\label{fig:real_robot_results}
\end{center}

\begin{figure*}[!t]
\centering
\includegraphics[width=0.98\textwidth]{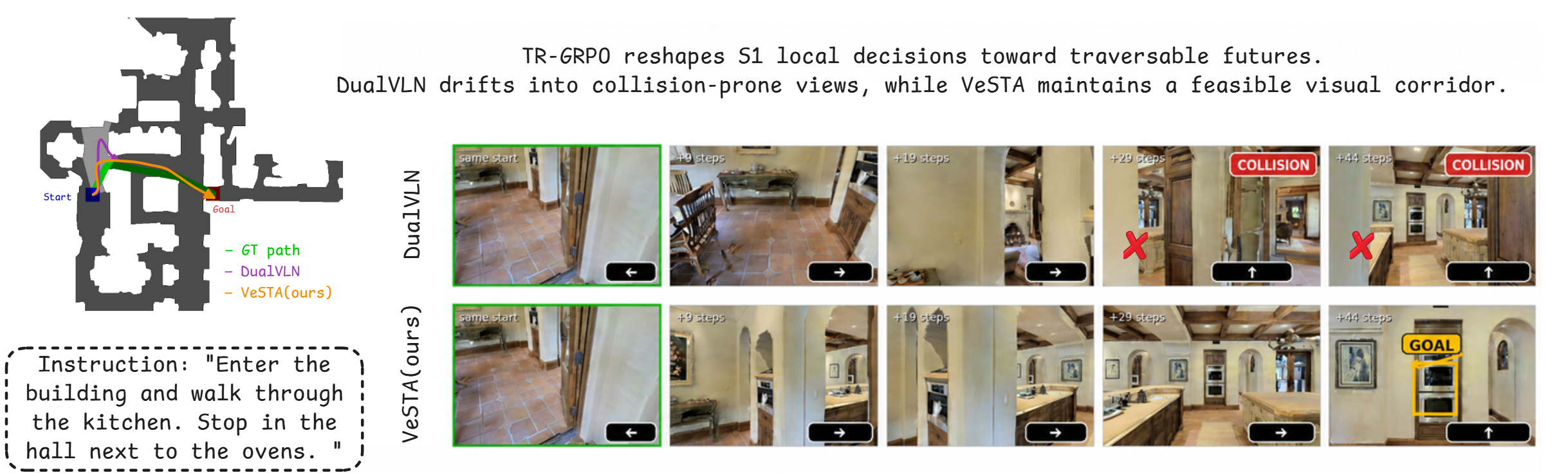}
\caption{Effect of Trajectory Relative Group Policy Optimization on the local trajectory distribution. Under the same instruction and scene, the original policy drifts into collision-prone views, whereas the aligned policy follows a geometrically feasible visual corridor, illustrating a shift toward the executable motion manifold.}
\label{fig:tr_grpo_decision_shift}
\end{figure*}

VeSTA increases observed CFSR from 80\% to 95\% in Hallway, 55\% to 65\% in Single Bedroom, and 30\% to 45\% in Room-to-Room (Figure~\ref{fig:real_robot_results}). Gains persist as geometric difficulty increases, although both policies decline. Under blocked contact, alignment favors visually plausible local futures that remain useful around furniture, door frames, and narrow passages.

\vspace{0.1em}
These trials should be interpreted as controlled evidence of applicability beyond simulation rather than a complete real-world safety evaluation. The number of scene groups is limited, the experiments are conducted in private indoor spaces without human participants in the navigation area, and CFSR does not characterize all physical failure modes. Nevertheless, the effect is consistent with strict-contact simulation: VeSTA improves the probability that semantic progress is realized without physical contact.

\vspace{-0.4em}
\subsection{Contact Semantics as a Controlled Intervention}
\vspace{-0.05em}
To reveal the physical executability gap, we hold each policy and episode specification fixed and intervene only on the contact operator. Table~\ref{tab:sliding} reports the paired permissive and strict conditions for StreamVLN, JanusVLN, and DualVLN.

\begin{table}[H]
\centering
\scriptsize
\setlength{\tabcolsep}{2.0pt}
\resizebox{\columnwidth}{!}{%
\begin{tabular}{llcccccc}
\toprule
Method & Contact Sem. & NE$\downarrow$ & OS$\uparrow$ & SR$\uparrow$ & SPL$\uparrow$ & Coll. Avg.$\downarrow$ & Coll. W.$\downarrow$ \\
\midrule
StreamVLN & permissive & 4.87 & 64.38 & 57.10 & 50.53 & 5.58 & 22.64 \\
StreamVLN & strict     & 6.19 & 37.74 & 30.30 & 27.59 & 39.03 & 67.85 \\
\midrule
JanusVLN & permissive & 4.96 & 63.30 & 58.02 & 54.53 & 8.17 & 25.04 \\
JanusVLN & strict     & 6.79 & 30.40 & 24.80 & 23.39 & 54.52 & 77.64 \\
\midrule
DualVLN & permissive & 4.03 & 70.96 & 65.31 & 59.08 & 2.47 & 9.03 \\
DualVLN & strict     & 4.84 & 57.64 & 51.50 & 47.61 & 13.22 & 34.00 \\
\bottomrule
\end{tabular}
}
\caption{Contact-semantics intervention on R2R Val-Unseen. Permissive contact can convert obstacle-directed motion into tangential displacement, whereas strict contact treats it as blocked execution. The performance difference between the two conditions diagnoses the physical executability gap. Collision metrics are percentages.}
\label{tab:sliding}
\end{table}

All three policies are sensitive to this intervention. StreamVLN loses 26.80 SR points and JanusVLN loses 33.22 points, while their weighted collision burdens rise to 67.85 and 77.64, respectively. DualVLN still loses 13.81 SR points as Coll. W. rises from 9.03 to 34.00. Because policy parameters and navigation episodes are unchanged, this contrast indicates system-level dependence of measured goal-reaching performance on permissive contact realization. It does not isolate the causal contribution of individual collisions, because changed contact outcomes also alter subsequent observations and episode evolution, nor does it imply that all permissive-condition success is spurious; rather, it quantifies sensitivity to the contact assumption when environment-side correction is removed.

VeSTA is evaluated in the strict condition precisely because this condition removes the compensatory mechanism. Its simultaneous improvement in SR and collision burden relative to strict-contact DualVLN indicates that the method recovers executable progress rather than merely exploiting a different contact response.

Permissive evaluation remains useful because it reflects a common operating assumption under which prior systems were developed, while strict contact provides a complementary diagnostic of embodied realization. Reporting both conditions reveals whether navigation conclusions remain stable when non-executable motion is no longer repaired.

\vspace{-0.4em}
\subsection{Mechanistic Ablation}
\vspace{-0.15em}
The two proposed mechanisms have different optimization roles: TR-GRPO changes candidate generation, whereas TRiC changes candidate selection. Table~\ref{tab:ablation_components} reports their component-level outcome effects under strict contact semantics; it does not directly measure executable-candidate coverage or selection regret. Together, the design predicts a sequential interaction in which candidate support and risk-aware selection contribute complementary reductions in collision burden.

TR-GRPO alone reduces Coll. W. from 34.00 to 29.71, showing that its training intervention can reduce contact burden, although it does not improve SR in isolation. TRiC reduces Coll. W. to 14.31 and exhibits a larger component-level effect under the present DualVLN instantiation. The full model achieves the lowest collision burden and highest SR/SPL, supporting outcome-level complementarity.

\begin{table}[H]
\centering
\small
\setlength{\tabcolsep}{2.3pt}
\begin{tabular}{lcccccc}
\toprule
Variant & NE$\downarrow$ & OS$\uparrow$ & SR$\uparrow$ & SPL$\uparrow$ & Coll. Avg.$\downarrow$ & Coll. W.$\downarrow$ \\
\midrule
DualVLN & 4.84 & 57.64 & 51.50 & 47.61 & 13.22 & 34.00 \\
+TR-GRPO & 4.88 & 56.82 & 50.84 & 46.61 & 12.23 & 29.71 \\
+TRiC & 4.64 & 59.65 & 52.15 & 47.29 & 7.03 & 14.31 \\
\textbf{VeSTA (Ours)} & \textbf{4.63} & \textbf{59.38} & \textbf{53.56} & \textbf{48.93} & \textbf{6.08} & \textbf{12.73} \\
\bottomrule
\end{tabular}
\caption{Component ablation under strict contact semantics. TR-GRPO shapes executable candidate support, TRiC performs candidate-level utility selection, and VeSTA combines both mechanisms. Collision metrics are percentages.}
\label{tab:ablation_components}
\end{table}

\FloatBarrier
\vspace{-0.6em}
\subsection{Decision-Level Evidence}
\vspace{-0.2em}
The aggregate ablation is complemented by a decision-level examination. TR-GRPO reshapes candidate support toward executable futures, whereas TRiC represents multi-factor strict-contact preferences through candidate-specific scalar utility.

\vspace{-0.35em}
\begin{center}
\includegraphics[width=1.0\columnwidth]{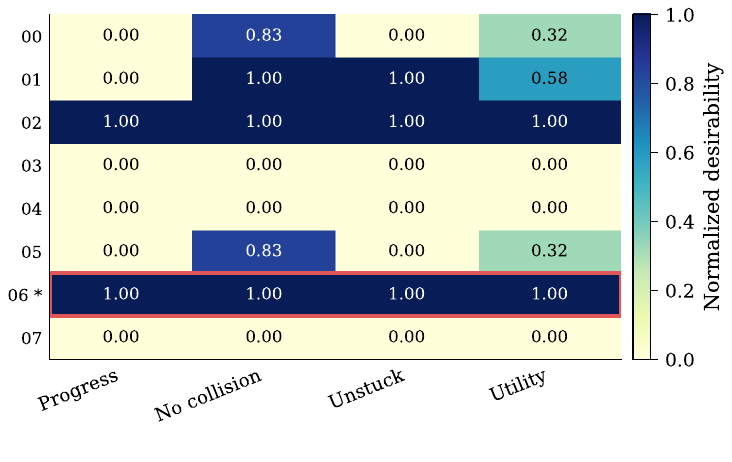}
\captionof{figure}{TRiC candidate selection. Each trajectory receives risk-adjusted utility from progress, contact, and immobilization preferences; the red box marks the selected candidate. Candidate-wise scoring preserves distinct motion modes.}
\label{fig:decision_heatmap}
\end{center}

\vspace{-0.3em}
Figure~\ref{fig:decision_heatmap} shows candidates trading progress against contact and immobilization risk. The selected trajectory maximizes risk-adjusted utility rather than progress alone or an average of modes. This example illustrates, rather than statistically proves, the mechanism in Equation~\ref{eq:tric_score}.

\section{Discussion}
\paragraph{Implications.} The results separate semantic intent from embodied realization. An agent may identify an appropriate target yet fail to execute the corresponding local motion; permissive dynamics can then make goal metrics reflect both policy competence and environment-side correction. Visual-semantic traversability connects these levels by requiring instruction consistency, geometric feasibility, and robustness to persistent contact or immobilization. This view treats collision not only as a safety event but also as evidence that an instruction-relevant local future was misjudged. TR-GRPO changes which futures are proposed, whereas TRiC selects among remaining modes. Their relative importance may vary with the quality and diversity of the base generator: a weak generator leaves little for selection to recover, while a diverse generator increases the value of candidate-specific risk reasoning. The ablation pattern shows a larger component-level effect for selection in the present baseline, with distribution alignment providing complementary gains in the combined model. More broadly, paired permissive and strict evaluation acts as a robustness test for contact assumptions. Reporting both absolute strict-contact performance and permissive-to-strict degradation distinguishes policies that execute progress from those whose apparent competence depends on environment-side repair.

\paragraph{Limitations.} Strict contact is a diagnostic abstraction, not a safety certificate; it omits compliance, slippage, actuator dynamics, recoverable contact, dynamic obstacles, social constraints, energy use, and severity-weighted risk. VeSTA cannot repair incorrect high-level subgoals, and its diagonal trajectory-vector ratio may be inaccurate for strongly non-Gaussian candidate sets. Aggregate tables lack repeated-seed confidence intervals; robot trials use 20 instructions per scene and CFSR conflates goal failure with contact. Component ablations do not directly measure candidate coverage or selection regret, and VeSTA is instantiated on one backbone. Thus large collision reductions are the primary evidence, while small navigation changes and cross-architecture rankings are descriptive. Non-decreasing SR/SPL rules out complete motion suppression but not subtler conservatism.

\paragraph{Future directions.} Future work should report confidence intervals, path length, executed steps, early stopping, no-progress rate, disaggregated physical outcomes, and support-versus-selection diagnostics. Calibrated trajectory-level risk and richer likelihoods could support adaptive replanning across embodiments, dynamic obstacles, and recoverable contact. RGB-only traversability should complement, not replace, specialized depth-, LiDAR-, or local-planner safety modules. Logging candidate feasibility before selection and realized outcomes after execution would further separate support, ranking, and controller failures.

\section{Conclusion}
Contact intervention shows that goal-reaching success need not imply physically executable motion. VeSTA combines trajectory distribution alignment with candidate-level risk scoring, substantially reducing collision-induced failures on R2R and RxR while preserving reported goal-reaching performance. Controlled Unitree Go2 trials provide evidence beyond simulation and suggest that strict-contact alignment captures a transferable preference for locally realizable motion. These results motivate continuous VLN evaluation that reports both semantic success and the execution reliability of the behavior used to achieve it. Operationally, contact semantics should therefore be treated as part of the benchmark specification rather than an implementation detail, with local executability reported alongside instruction completion.

\bibliographystyle{plainnat}
\bibliography{references}

@inproceedings{anderson2018vision,
  title = {Vision-and-Language Navigation: Interpreting Visually-Grounded Navigation Instructions in Real Environments},
  author = {Anderson, Peter and Wu, Qi and Teney, Damien and Bruce, Jake and Johnson, Mark and S{\"u}nderhauf, Niko and Reid, Ian and Gould, Stephen and van den Hengel, Anton},
  booktitle = {Proceedings of the IEEE Conference on Computer Vision and Pattern Recognition},
  pages = {3674--3683},
  year = {2018}
}

@inproceedings{krantz2020beyond,
  title = {Beyond the Nav-Graph: Vision-and-Language Navigation in Continuous Environments},
  author = {Krantz, Jacob and Wijmans, Erik and Majumdar, Arjun and Batra, Dhruv and Lee, Stefan},
  booktitle = {Proceedings of the European Conference on Computer Vision},
  pages = {104--120},
  year = {2020},
  organization = {Springer}
}

@inproceedings{savva2019habitat,
  title = {Habitat: A Platform for Embodied AI Research},
  author = {Savva, Manolis and Kadian, Abhishek and Maksymets, Oleksandr and Zhao, Yili and Wijmans, Erik and Jain, Bhavana and Straub, Julian and Liu, Jia and Koltun, Vladlen and Malik, Jitendra and Parikh, Devi and Batra, Dhruv},
  booktitle = {Proceedings of the IEEE/CVF International Conference on Computer Vision},
  pages = {9339--9347},
  year = {2019}
}

@misc{streamvln2025,
  title = {StreamVLN: Streaming Vision-and-Language Navigation via SlowFast Context Modeling},
  author = {Wei, Meng and Wan, Chenyang and Yu, Xiqian and Wang, Tai and Yang, Yuqiang and Mao, Xiaohan and Zhu, Chenming and Cai, Wenzhe and Wang, Hanqing and Chen, Yilun and Liu, Xihui and Pang, Jiangmiao},
  year = {2025},
  eprint = {2507.05240},
  archivePrefix = {arXiv},
  primaryClass = {cs.CV}
}

@misc{janusvln2025,
  title = {JanusVLN: Decoupling Semantics and Spatiality with Dual Implicit Memory for Vision-Language Navigation},
  author = {Zeng, Shuang and Qi, Dekang and Chang, Xinyuan and Xiong, Feng and Xie, Shichao and Wu, Xiaolong and Liang, Shiyi and Xu, Mu and Wei, Xing and Guo, Ning},
  year = {2025},
  eprint = {2509.22548},
  archivePrefix = {arXiv},
  primaryClass = {cs.CV}
}

@misc{dualvln2025,
  title = {Ground Slow, Move Fast: A Dual-System Foundation Model for Generalizable Vision-and-Language Navigation},
  author = {Wei, Meng and Wan, Chenyang and Peng, Jiaqi and Yu, Xiqian and Yang, Yuqiang and Feng, Delin and Cai, Wenzhe and Zhu, Chenming and Wang, Tai and Pang, Jiangmiao and Liu, Xihui},
  year = {2025},
  eprint = {2512.08186},
  archivePrefix = {arXiv},
  primaryClass = {cs.CV}
}

@misc{shao2024deepseekmath,
  title = {DeepSeekMath: Pushing the Limits of Mathematical Reasoning in Open Language Models},
  author = {Shao, Zhihong and Wang, Peiyi and Zhu, Qihao and Xu, Runxin and Song, Junxiao and Bi, Xiao and Zhang, Haowei and Zhang, Mingchuan and Li, Y. K. and Wu, Y. and Guo, Daya},
  year = {2024},
  eprint = {2402.03300},
  archivePrefix = {arXiv},
  primaryClass = {cs.CL}
}

@article{fox1997dynamic,
  title = {The Dynamic Window Approach to Collision Avoidance},
  author = {Fox, Dieter and Burgard, Wolfram and Thrun, Sebastian},
  journal = {IEEE Robotics \& Automation Magazine},
  volume = {4},
  number = {1},
  pages = {23--33},
  year = {1997},
  doi = {10.1109/100.580977}
}

@inproceedings{rosmann2013efficient,
  title = {Efficient Trajectory Optimization Using a Sparse Model},
  author = {R{\"o}smann, Christoph and Feiten, Wendelin and W{\"o}sch, Thomas and Hoffmann, Frank and Bertram, Torsten},
  booktitle = {Proceedings of the European Conference on Mobile Robots},
  pages = {138--143},
  year = {2013},
  doi = {10.1109/ECMR.2013.6698833}
}

@misc{yue2023safevln,
  title = {Safe-VLN: Collision Avoidance for Vision-and-Language Navigation of Autonomous Robots Operating in Continuous Environments},
  author = {Yue, Lu and Zhou, Dongliang and Xie, Liang and Zhang, Feitian and Yan, Ye and Yin, Erwei},
  year = {2023},
  eprint = {2311.02817},
  archivePrefix = {arXiv},
  primaryClass = {cs.RO}
}

@misc{jeong2024vlncm,
  title = {Zero-Shot Vision-and-Language Navigation with Collision Mitigation in Continuous Environment},
  author = {Jeong, Seongjun and Kang, Gi-Cheon and Kim, Joochan and Zhang, Byoung-Tak},
  year = {2024},
  eprint = {2410.17267},
  archivePrefix = {arXiv},
  primaryClass = {cs.RO}
}

@inproceedings{kim2025care,
  title = {Enhancing Safety of Foundation Models for Visual Navigation through Collision Avoidance via Repulsive Estimation},
  author = {Kim, Joonkyung and Sim, Joonyeol and Kim, Woojun and Sycara, Katia and Nam, Changjoo},
  booktitle = {Proceedings of The 9th Conference on Robot Learning},
  pages = {3704--3719},
  year = {2025}
}

@misc{zhang2024navid,
  title = {NaVid: Video-based VLM Plans the Next Step for Vision-and-Language Navigation},
  author = {Zhang, Jiazhao and Wang, Kunyu and Xu, Rongtao and Zhou, Gengze and Hong, Yicong and Fang, Xiaomeng and Wu, Qi and Zhang, Zhizheng and Wang, He},
  year = {2024},
  eprint = {2402.15852},
  archivePrefix = {arXiv},
  primaryClass = {cs.CV}
}

@misc{cheng2024navila,
  title = {NaVILA: Legged Robot Vision-Language-Action Model for Navigation},
  author = {Cheng, An-Chieh and Ji, Yandong and Yang, Zhaojing and Gongye, Zaitian and Zou, Xueyan and Kautz, Jan and Biyik, Erdem and Yin, Hongxu and Liu, Sifei and Wang, Xiaolong},
  year = {2024},
  eprint = {2412.04453},
  archivePrefix = {arXiv},
  primaryClass = {cs.RO}
}

@misc{zhang2024uninavid,
  title = {Uni-NaVid: A Video-based Vision-Language-Action Model for Unifying Embodied Navigation Tasks},
  author = {Zhang, Jiazhao and Wang, Kunyu and Wang, Shaoan and Li, Minghan and Liu, Haoran and Zhou, Gengze and Li, Yicong and Fang, Xiaomeng and Wu, Qi and Zhang, Zhizheng and Wang, He},
  year = {2024},
  eprint = {2412.06224},
  archivePrefix = {arXiv},
  primaryClass = {cs.CV}
}

@misc{schulman2017ppo,
  title = {Proximal Policy Optimization Algorithms},
  author = {Schulman, John and Wolski, Filip and Dhariwal, Prafulla and Radford, Alec and Klimov, Oleg},
  year = {2017},
  eprint = {1707.06347},
  archivePrefix = {arXiv},
  primaryClass = {cs.LG}
}

@article{anderson2018evaluation,
  title = {On Evaluation of Embodied Navigation Agents},
  author = {Anderson, Peter and Chang, Angel and Chaplot, Devendra Singh and Dosovitskiy, Alexey and Gupta, Saurabh and Koltun, Vladlen and Kosecka, Jana and Malik, Jitendra and Mottaghi, Roozbeh and Savva, Manolis and Zamir, Amir R.},
  journal = {arXiv preprint arXiv:1807.06757},
  year = {2018},
  eprint = {1807.06757},
  archivePrefix = {arXiv},
  primaryClass = {cs.AI}
}

@misc{lyu2025fpo,
  title = {Reinforcement Fine-Tuning of Flow-Matching Policies for Vision-Language-Action Models},
  author = {Lyu, Mingyang and Sun, Yinqian and Lin, Erliang and Li, Huangrui and Chen, Ruolin and Zhao, Feifei and Zeng, Yi},
  year = {2025},
  eprint = {2510.09976},
  archivePrefix = {arXiv},
  primaryClass = {cs.LG}
}

@misc{unitreego2,
  title = {Unitree Go2: Quadruped Robot Platform},
  author = {{Unitree Robotics}},
  year = {2024},
  howpublished = {\url{https://www.unitree.com/go2}},
  note = {Accessed 2026-06-27}
}

@inproceedings{ku2020rxr,
  title = {Room-Across-Room: Multilingual Vision-and-Language Navigation with Dense Spatiotemporal Grounding},
  author = {Ku, Alexander and Anderson, Peter and Patel, Roma and Ie, Eugene and Baldridge, Jason},
  booktitle = {Proceedings of the 2020 Conference on Empirical Methods in Natural Language Processing},
  pages = {4392--4412},
  year = {2020},
  publisher = {Association for Computational Linguistics},
  doi = {10.18653/v1/2020.emnlp-main.356}
}

@article{kadian2020sim2real,
  title = {Sim2Real Predictivity: Does Evaluation in Simulation Predict Real-World Performance?},
  author = {Kadian, Abhishek and Truong, Joanne and Gokaslan, Aaron and Clegg, Alexander and Wijmans, Erik and Lee, Stefan and Savva, Manolis and Chernova, Sonia and Batra, Dhruv},
  journal = {IEEE Robotics and Automation Letters},
  volume = {5},
  number = {4},
  pages = {6670--6677},
  year = {2020},
  doi = {10.1109/LRA.2020.3013848}
}

@inproceedings{krantz2021waypoint,
  title = {Waypoint Models for Instruction-Guided Navigation in Continuous Environments},
  author = {Krantz, Jacob and Gokaslan, Aaron and Batra, Dhruv and Lee, Stefan and Maksymets, Oleksandr},
  booktitle = {Proceedings of the IEEE/CVF International Conference on Computer Vision},
  pages = {15162--15171},
  year = {2021}
}

@inproceedings{hong2022bridging,
  title = {Bridging the Gap Between Learning in Discrete and Continuous Environments for Vision-and-Language Navigation},
  author = {Hong, Yicong and Wang, Zun and Wu, Qi and Gould, Stephen},
  booktitle = {Proceedings of the IEEE/CVF Conference on Computer Vision and Pattern Recognition},
  pages = {15439--15449},
  year = {2022}
}

@inproceedings{georgakis2022crossmodal,
  title = {Cross-Modal Map Learning for Vision and Language Navigation},
  author = {Georgakis, Georgios and Schmeckpeper, Karl and Wanchoo, Karan and Dan, Soham and Miltsakaki, Eleni and Roth, Dan and Daniilidis, Kostas},
  booktitle = {Proceedings of the IEEE/CVF Conference on Computer Vision and Pattern Recognition},
  pages = {15460--15470},
  year = {2022}
}

@inproceedings{wang2025embodiedgap,
  title = {Rethinking the Embodied Gap in Vision-and-Language Navigation: A Holistic Study of Physical and Visual Disparities},
  author = {Wang, Liuyi and Xia, Xinyuan and Zhao, Hui and Wang, Hanqing and Wang, Tai and Chen, Yilun and Liu, Chengju and Chen, Qijun and Pang, Jiangmiao},
  booktitle = {Proceedings of the IEEE/CVF International Conference on Computer Vision},
  pages = {9455--9465},
  year = {2025}
}

@misc{ilharco2019dtw,
  title = {General Evaluation for Instruction Conditioned Navigation using Dynamic Time Warping},
  author = {Ilharco, Gabriel and Jain, Vihan and Ku, Alexander and Ie, Eugene and Baldridge, Jason},
  year = {2019},
  eprint = {1907.05446},
  archivePrefix = {arXiv},
  primaryClass = {cs.CL}
}

@article{an2025etpnav,
  title = {ETPNav: Evolving Topological Planning for Vision-Language Navigation in Continuous Environments},
  author = {An, Dong and Wang, Hanqing and Wang, Wenguan and Wang, Zun and Huang, Yan and He, Keji and Wang, Liang},
  journal = {IEEE Transactions on Pattern Analysis and Machine Intelligence},
  volume = {47},
  number = {7},
  pages = {5130--5145},
  year = {2025},
  doi = {10.1109/TPAMI.2025.3565877}
}

@inproceedings{lipman2023flow,
  title = {Flow Matching for Generative Modeling},
  author = {Lipman, Yaron and Chen, Ricky T. Q. and Ben-Hamu, Heli and Nickel, Maximilian and Le, Matt},
  booktitle = {International Conference on Learning Representations},
  year = {2023}
}

@inproceedings{li2026navgrpo,
  title = {NavGRPO: Trajectory-Diversity-Driven Robust Vision-and-Language Navigation},
  author = {Li, Xiang and Zhang, Zhong-Yu and Guan, Yifei and Wu, Quan and Cheng, Yajing and Yan, Zhiwei},
  booktitle = {Proceedings of the IEEE/CVF Conference on Computer Vision and Pattern Recognition},
  pages = {6044--6054},
  year = {2026}
}

@inproceedings{lee2017desire,
  title = {DESIRE: Distant Future Prediction in Dynamic Scenes with Interacting Agents},
  author = {Lee, Namhoon and Choi, Wongun and Vernaza, Paul and Choy, Christopher B. and Torr, Philip H. S. and Chandraker, Manmohan},
  booktitle = {Proceedings of the IEEE Conference on Computer Vision and Pattern Recognition},
  pages = {336--345},
  year = {2017}
}

@inproceedings{chai2020multipath,
  title = {MultiPath: Multiple Probabilistic Anchor Trajectory Hypotheses for Behavior Prediction},
  author = {Chai, Yuning and Sapp, Benjamin and Bansal, Mayank and Anguelov, Dragomir},
  booktitle = {Proceedings of the Conference on Robot Learning},
  series = {Proceedings of Machine Learning Research},
  volume = {100},
  pages = {86--99},
  year = {2020}
}

@inproceedings{chen2024mapgpt,
  title = {MapGPT: Map-Guided Prompting for Unified Vision-and-Language Navigation},
  author = {Chen, Jiaqi and Li, Bingqian and Xu, Xiangyang and Tu, Zhaoxiang and Xiang, Hongbo and Wang, Cen and He, Hongbo},
  booktitle = {Proceedings of the 62nd Annual Meeting of the Association for Computational Linguistics},
  pages = {9796--9810},
  year = {2024},
  publisher = {Association for Computational Linguistics},
  doi = {10.18653/v1/2024.acl-long.529}
}

@inproceedings{zhou2024navgpt2,
  title = {NavGPT-2: Unleashing Navigational Reasoning Capability for Large Vision-Language Models},
  author = {Zhou, Gengze and Hong, Yicong and Wang, Zun and Wang, Xin Eric and Wu, Qi},
  booktitle = {Proceedings of the European Conference on Computer Vision},
  pages = {260--278},
  year = {2024},
  organization = {Springer}
}

@inproceedings{shah2023vint,
  title = {ViNT: A Foundation Model for Visual Navigation},
  author = {Shah, Dhruv and Sridhar, Ajay and Bhorkar, Arjun and Hirose, Noriaki and Levine, Sergey},
  booktitle = {Proceedings of the Conference on Robot Learning},
  series = {Proceedings of Machine Learning Research},
  volume = {229},
  pages = {404--418},
  year = {2023}
}

@inproceedings{sridhar2024nomad,
  title = {NoMaD: Goal Masked Diffusion Policies for Navigation and Exploration},
  author = {Sridhar, Ajay and Shah, Dhruv and Levine, Sergey and Kumar, Sudeep},
  booktitle = {Proceedings of the IEEE International Conference on Robotics and Automation},
  pages = {63--70},
  year = {2024},
  doi = {10.1109/ICRA57147.2024.10610796}
}

@inproceedings{chi2023diffusion,
  title = {Diffusion Policy: Visuomotor Policy Learning via Action Diffusion},
  author = {Chi, Cheng and Feng, Siyuan and Du, Yilun and Xu, Zhenjia and Cousineau, Eric and Burchfiel, Benjamin and Song, Shuran},
  booktitle = {Robotics: Science and Systems},
  year = {2023},
  doi = {10.15607/RSS.2023.XIX.026}
}

@inproceedings{an2023bevbert,
  title = {BEVBert: Multimodal Map Pre-training for Language-guided Navigation},
  author = {An, Dong and Qi, Yuankai and Li, Yangguang and Huang, Yan and Wang, Liang and Tan, Tieniu and Shao, Jing},
  booktitle = {Proceedings of the IEEE/CVF International Conference on Computer Vision},
  pages = {10813--10823},
  year = {2023}
}

@inproceedings{wang2023gridmm,
  title = {GridMM: Grid Memory Map for Vision-and-Language Navigation},
  author = {Wang, Zihan and Li, Xiangyang and Yang, Jiahao and Liu, Yeqi and Jiang, Shuqiang},
  booktitle = {Proceedings of the IEEE/CVF International Conference on Computer Vision},
  pages = {15625--15636},
  year = {2023}
}

@inproceedings{wang2023dreamwalker,
  title = {DREAMWALKER: Mental Planning for Continuous Vision-Language Navigation},
  author = {Wang, Hanqing and Liang, Wei and Van Gool, Luc and Wang, Wenguan},
  booktitle = {Proceedings of the IEEE/CVF International Conference on Computer Vision},
  pages = {10873--10883},
  year = {2023}
}

@inproceedings{wang2023scalevln,
  title = {Scaling Data Generation in Vision-and-Language Navigation},
  author = {Wang, Hanqing and Liang, Wei and Van Gool, Luc and Wang, Wenguan},
  booktitle = {Proceedings of the IEEE/CVF International Conference on Computer Vision},
  pages = {12009--12020},
  year = {2023}
}

@inproceedings{guo2026awarevln,
  title = {AwareVLN: Vision-Language Navigation with Dynamic Area Awareness and Self-Aware Reasoning},
  author = {Guo, Chen and Zhang, Hanming and Li, Shuang and Liu, Zhaoyang and Chen, Chao and Li, Wenguan and Zhang, Shurui and Fu, Sihang and Pang, Yijin and Shen, Xiaochao},
  booktitle = {Proceedings of the IEEE/CVF Conference on Computer Vision and Pattern Recognition},
  pages = {9381--9391},
  year = {2026}
}

@inproceedings{dai2026evonav,
  title = {History to Future: Evolving Agent with Experience and Thought for Zero-Shot Vision-and-Language Navigation},
  author = {Dai, Guangzhao and Wang, Shuo and Wang, Zihan and Xie, Guo-Sen and Yang, Yang and Pan, Jinshan and Sun, Qianru and Shu, Xiangbo},
  booktitle = {Proceedings of the IEEE/CVF Conference on Computer Vision and Pattern Recognition},
  pages = {15177--15187},
  year = {2026}
}

@inproceedings{pashevich2021topological,
  title = {Topological Planning with Transformers for Vision-and-Language Navigation},
  author = {Pashevich, Alexander and Schmid, Cordelia and Sun, Chen and Lopes, Pedro H. O. Pinheiro and Sener, Ozan},
  booktitle = {Proceedings of the IEEE/CVF International Conference on Computer Vision},
  pages = {11276--11286},
  year = {2021}
}

@inproceedings{ghaus2025navmorph,
  title = {NavMorph: A Self-Evolving World Model for Vision-Language Navigation in Continuous Environments},
  author = {Ghaus, Saad and Ivanovic, Boris and Pavone, Marco},
  booktitle = {Proceedings of the IEEE/CVF International Conference on Computer Vision},
  pages = {15106--15116},
  year = {2025}
}

\clearpage
\onecolumn
\appendix
\renewcommand{\thesection}{A\arabic{section}}
\renewcommand{\thesubsection}{A\arabic{section}.\arabic{subsection}}
\setcounter{table}{0}
\renewcommand{\thetable}{A\arabic{table}}
\setlength{\parindent}{1.5em}
\renewcommand{\arraystretch}{1.12}
\section*{Appendix: Additional Analyses}
\setcounter{section}{0}

\section{Scope of the Appendix}
This appendix consolidates statistical and mechanism-oriented analyses that support the claims in the main paper without expanding the primary claim scope. The analyses use existing episode-level evaluation logs and robot outcome counts; they do not involve additional large-scale training. We examine paired uncertainty, candidate support and selection, component interaction, trajectory length, collision-tail behavior, baseline-risk heterogeneity, critic ranking quality, and uncertainty in the small physical study. We also state explicitly which mechanism-level conclusions are not established by the available evidence.

\section{Operational Clarifications}
\subsection{Training supervision and inference boundary}
Strict-contact state restoration, collision events, clearance estimates, progress, and immobilization indicators are used only to construct training or diagnostic preferences. At inference, VeSTA consumes the same RGB observation and instruction-conditioned subgoal interface as its DualVLN backbone. It does not access simulator state, privileged geometry, depth, LiDAR, or an external collision sensor or local planner. Thus, the additional supervision is privileged during training, whereas deployment remains RGB-only.

\subsection{Interpretation of TR-GRPO}
The reported TR-GRPO configuration combines the standard PPO actor--critic objective with a clipped group-relative trajectory term. Candidate trajectories are represented as vectors in $\mathbb{R}^{3H}$ with $H=6$, and two reparameterized candidates are probed by executing their first four steps from an identical restored state. Because the reported group size is $G=2$, the normalized group advantages are approximately opposite-signed; the implemented configuration is therefore most precisely interpreted as \emph{pairwise group-relative trajectory optimization}. The current component result identifies the effect of the complete TR-GRPO training intervention, rather than isolating the group-relative term from PPO fine-tuning and reward shaping.

\subsection{Unsafe-candidate penalty}
The additional penalty assigned to collision, clearance below $0.15$ m, or non-progress is intentionally much larger than ordinary reward terms. It implements an approximately lexicographic preference in which clearly non-executable candidates are rejected before finer progress--clearance trade-offs are considered. This design should not be interpreted as a calibrated physical risk model. The present evidence evaluates the complete reward configuration and does not isolate sensitivity to the penalty magnitude.

\section{Statistical Protocol}
All simulation uncertainty analyses use paired episode bootstrap with $10{,}000$ resamples and fixed seed 20260727. Episodes are aligned by scene and episode identifier, and each resample draws aligned episode pairs with replacement. Reported intervals are percentile 95\% confidence intervals for the method-minus-baseline difference. SR, SPL, OS, CollAvg, and nDTW are episode averages. CollW is recomputed within every resample as the ratio of total contacts to total executed steps, rather than bootstrapping a pre-aggregated scalar.

R2R comparisons contain 1,839 paired Val-Unseen episodes. RxR comparisons contain 3,669 paired English Val-Unseen episodes. Following the cleaned reporting convention used by the main paper, non-finite RxR SPL values are treated as zero. Ninety-nine episodes from the same scene contain non-finite NE values for both methods; RxR NE is therefore averaged over the remaining 3,570 finite paired episodes. Bootstrap intervals characterize episode-level sampling uncertainty for the evaluated checkpoints; they do not replace repeated-training-seed uncertainty.

\section{Paired Uncertainty Analysis}
\subsection{R2R component comparisons}
Table~\ref{tab:r2r_bootstrap} reports paired differences relative to strict-contact DualVLN. Collision changes are consistently negative. The TR-GRPO intervention has a smaller but nonzero collision effect: CollW decreases by 4.29 percentage points with a 95\% interval of $[-6.00,-2.61]$. TRiC and the complete VeSTA model produce substantially larger collision reductions. In contrast, most SR and SPL intervals overlap zero. This supports treating collision reduction as the primary result and small navigation changes as descriptive, as done in the main paper.

\begin{table}[t]
\centering
\small
\setlength{\tabcolsep}{8pt}
\begin{tabular}{lcc}
\toprule
Comparison & $\Delta$SR (95\% CI) & $\Delta$SPL (95\% CI) \\
\midrule
+TR-GRPO vs. DualVLN & $-0.65$ [$-2.56,1.31$] & $-1.00$ [$-2.73,0.74$] \\
+TRiC vs. DualVLN & $+0.65$ [$-1.52,2.83$] & $-0.32$ [$-2.23,1.64$] \\
VeSTA vs. DualVLN & $+2.07$ [$-0.11,4.19$] & $+1.33$ [$-0.63,3.27$] \\
\midrule
Comparison & $\Delta$CollAvg (95\% CI) & $\Delta$CollW (95\% CI) \\
\midrule
+TR-GRPO vs. DualVLN & $-0.99$ [$-1.93,-0.06$] & $-4.29$ [$-6.00,-2.61$] \\
+TRiC vs. DualVLN & $-6.20$ [$-7.24,-5.19$] & $-19.69$ [$-21.50,-17.85$] \\
VeSTA vs. DualVLN & $-7.14$ [$-8.17,-6.13$] & $-21.27$ [$-23.12,-19.42$] \\
\bottomrule
\end{tabular}
\caption{Paired episode bootstrap on R2R Val-Unseen under strict contact. The upper panel reports navigation outcomes and the lower panel reports collision outcomes. Values are percentage-point method-minus-baseline differences.}
\label{tab:r2r_bootstrap}
\end{table}

The full model reduces CollAvg from 13.22\% to 6.08\%, a 54.0\% relative reduction, and CollW from 34.00\% to 12.73\%, a 62.6\% relative reduction. TRiC also improves NE by $-0.199$ m with a 95\% interval of $[-0.390,-0.010]$, while VeSTA improves NE by $-0.206$ m with an interval of $[-0.388,-0.027]$. These secondary results are compatible with preserved navigation but are smaller than the collision effects.

\subsection{RxR transfer}
The RxR comparison follows the same pattern (Table~\ref{tab:rxr_bootstrap}). VeSTA reduces CollAvg by 12.44 points and CollW by 23.64 points, with both intervals well separated from zero. Relative reductions are 57.0\% and 60.4\%, respectively. The SR and SPL intervals overlap zero, whereas finite-only NE and nDTW exhibit smaller favorable differences. Consequently, the cross-dataset evidence supports robust collision reduction more strongly than a general claim of improved goal-reaching success.

\begin{table}[t]
\centering
\small
\setlength{\tabcolsep}{6pt}
\begin{tabular}{lcc}
\toprule
Metric & Difference & Paired bootstrap 95\% CI \\
\midrule
SR & $+0.87$ & $[-0.71,2.43]$ \\
SPL & $+0.52$ & $[-0.82,1.81]$ \\
NE & $-0.304$ & $[-0.496,-0.116]$ \\
nDTW & $+1.32$ & $[0.48,2.18]$ \\
CollAvg & $-12.44$ & $[-13.24,-11.62]$ \\
CollW & $-23.64$ & $[-24.63,-22.62]$ \\
\bottomrule
\end{tabular}
\caption{VeSTA-minus-DualVLN paired bootstrap on the RxR English Val-Unseen subset. SR, SPL, nDTW, and collision values are percentage-point differences; NE is measured in meters. NE uses 3,570 finite paired episodes, while the other metrics use all 3,669 episodes.}
\label{tab:rxr_bootstrap}
\end{table}

\section{Instruction-Conditioned Robustness}
We further partition R2R episodes using instruction text only. Doorway-related instructions contain \emph{door}, \emph{arch}, \emph{entrance}, or \emph{exit}; long instructions contain more than 30 words. These groups are descriptive and are not treated as causal categories. Table~\ref{tab:instruction_groups} shows that collision reduction persists in every group. The subgroup intervals are exploratory, overlap across non-independent partitions, and are not corrected for multiple comparisons. Although the interval for instructions containing \emph{straight} does not cross zero, it is not treated as confirmatory evidence. The collision effect is not confined to one narrow linguistic pattern, while navigation changes remain heterogeneous.

\begin{table}[t]
\centering
\small
\setlength{\tabcolsep}{4.5pt}
\begin{tabular}{lrrcc}
\toprule
R2R group & Episodes & Share & $\Delta$SR (95\% CI) & $\Delta$CollAvg (95\% CI) \\
\midrule
Doorway terms present & 1,130 & 61.4\% & $+1.95$ [$-0.80,4.60$] & $-6.99$ [$-8.30,-5.69$] \\
Doorway terms absent & 709 & 38.6\% & $+2.26$ [$-1.27,5.92$] & $-7.39$ [$-9.07,-5.79$] \\
Instruction $>30$ words & 558 & 30.3\% & $+2.87$ [$-1.08,6.81$] & $-7.25$ [$-9.06,-5.44$] \\
Instruction $\leq30$ words & 1,281 & 69.7\% & $+1.72$ [$-0.78,4.22$] & $-7.10$ [$-8.29,-5.89$] \\
Contains \emph{straight} & 413 & 22.5\% & $+4.84$ [$0.48,9.20$] & $-4.81$ [$-6.87,-2.75$] \\
Does not contain \emph{straight} & 1,426 & 77.5\% & $+1.26$ [$-1.26,3.79$] & $-7.82$ [$-8.99,-6.65$] \\
\bottomrule
\end{tabular}
\caption{VeSTA-minus-DualVLN paired differences by instruction-derived R2R groups. Values are percentage points. Categories are overlapping descriptive partitions, and intervals are exploratory without multiple-comparison correction.}
\label{tab:instruction_groups}
\end{table}

\section{TRiC Risk Stratification}
The available TRiC logs contain selector-side collision-risk and stuck-risk scores for all 419 episodes recorded in the rank-0 evaluation shard. The analysis includes every record in that shard without outcome-based filtering, but the shard is not assumed to be a random sample of the complete evaluation set. We divide episodes into quartiles according to each risk score and report realized episode outcomes. This analysis tests risk stratification rather than probability calibration or candidate-selection optimality.

\begin{table}[t]
\centering
\small
\setlength{\tabcolsep}{4pt}
\begin{tabular}{lccc}
\toprule
Risk signal quartile & Episodes & SR & CollAvg \\
\midrule
Collision-risk score Q1 (lowest) & 105 & 84.76 & 0.70 \\
Collision-risk score Q2 & 104 & 74.04 & 3.69 \\
Collision-risk score Q3 & 105 & 49.52 & 3.87 \\
Collision-risk score Q4 (highest) & 105 & 22.86 & 19.19 \\
\midrule
Stuck-risk score Q1 (lowest) & 105 & 90.48 & 3.31 \\
Stuck-risk score Q2 & 104 & 80.77 & 3.91 \\
Stuck-risk score Q3 & 105 & 46.67 & 2.00 \\
Stuck-risk score Q4 (highest) & 105 & 13.33 & 18.23 \\
\bottomrule
\end{tabular}
\caption{Realized outcomes across quartiles of TRiC risk scores on the complete available rank-0 shard (419 episodes). Metrics are percentages. Endpoint separation supports descriptive risk stratification, but does not establish calibrated probabilities, causal effects, or optimal candidate selection.}
\label{tab:risk_quartiles}
\end{table}

The highest collision-risk quartile has 19.19\% realized CollAvg compared with 0.70\% in the lowest quartile. Likewise, SR decreases from 90.48\% to 13.33\% between the lowest and highest stuck-risk quartiles. The middle collision values are not perfectly monotonic, so these outputs should be used as descriptive ranking signals rather than interpreted as calibrated risk probabilities. Because no counterfactual candidate utilities are available in this shard, the analysis does not measure oracle utility or selection regret.

\section{Physical-Study Uncertainty}
The robot study contains 20 instructions per scene and method. Table~\ref{tab:robot_intervals} reports raw collision-free success counts and Wilson 95\% intervals. VeSTA improves the observed CFSR in every scene and in the pooled descriptive summary. However, scene-wise and pooled difference intervals remain wide and overlap zero under an unpaired normal approximation. The physical study therefore provides directional transfer evidence rather than a precise estimate of effect size.

\begin{table}[t]
\centering
\small
\setlength{\tabcolsep}{9pt}
\begin{tabular}{lcc}
\toprule
Scene & \shortstack{DualVLN\\count, CFSR [Wilson 95\% CI]} & \shortstack{VeSTA\\count, CFSR [Wilson 95\% CI]} \\
\midrule
Hallway & \shortstack{16/20, 80.0\\$[58.4,91.9]$} & \shortstack{19/20, 95.0\\$[76.4,99.1]$} \\
Bedroom & \shortstack{11/20, 55.0\\$[34.2,74.2]$} & \shortstack{13/20, 65.0\\$[43.3,81.9]$} \\
Room-to-Room & \shortstack{6/20, 30.0\\$[14.5,51.9]$} & \shortstack{9/20, 45.0\\$[25.8,65.8]$} \\
Pooled & \shortstack{33/60, 55.0\\$[42.5,66.9]$} & \shortstack{41/60, 68.3\\$[55.8,78.7]$} \\
\midrule
Scene & $\Delta$CFSR & Approx. 95\% CI for difference \\
\midrule
Hallway & $+15.0$ & $[-5.0,35.0]$ \\
Bedroom & $+10.0$ & $[-20.2,40.2]$ \\
Room-to-Room & $+15.0$ & $[-14.6,44.6]$ \\
Pooled & $+13.3$ & $[-3.9,30.6]$ \\
\bottomrule
\end{tabular}
\caption{Raw collision-free success counts, Wilson intervals for each method, and unpaired normal-approximation intervals for method differences. Percentages and differences are percentage points. The pooled row ignores scene heterogeneity and is descriptive.}
\label{tab:robot_intervals}
\end{table}

CFSR combines goal failure and contact failure. The current raw records do not contain a complete disaggregation into SR and contact-free rate, so no such values are inferred. A future release should record the two binary outcomes separately for every trial and report paired instruction-level differences when the same instructions are used for both methods.

\section{Historical Candidate Support and Selection}
We use the logged probe subset from \texttt{DualVLN\_260319\_selector\_reuse} to separate candidate support from candidate selection. A candidate is defined as executable when its local probe is collision-free, is not marked stuck, and achieves positive progress. The subset contains 78,979 decision points from 1,835 episodes. At least one executable representative is present at 37.64\% of decisions. The selected representative is executable at 34.86\% of all decisions and at 92.63\% of decisions with executable support.

On supported decisions, the selected candidate achieves mean local progress 0.707, compared with 0.767 for the progress-best executable candidate, leaving a mean gap of 0.059. When exact progress ties are counted as multiple valid best candidates, the selection matches a progress-best executable candidate at 79.31\% of supported decisions. Restricting to the 23,070 decisions with a unique progress-best executable candidate gives a 77.88\% match rate. These statistics show that selection is usually effective once the candidate set contains a locally executable option, while support remains absent at many decisions.

This diagnostic is historical rather than a final-checkpoint counterfactual evaluation. It therefore does not establish that TR-GRPO increased executable-candidate coverage relative to the baseline. Moreover, the progress-best candidate is only a local progress oracle, not an oracle for the complete risk-adjusted utility used by TRiC; the reported gap should not be interpreted as full selection regret.

\section{Component Interaction}
The component outcomes also clarify the meaning of ``complementary.'' For each collision metric, we compare the gain of each individual component with the gain of the full model, where a positive gain denotes a reduction from DualVLN.

\begin{table}[H]
\centering
\small
\setlength{\tabcolsep}{7pt}
\begin{tabular}{lrrrrr}
\toprule
Metric & TR-GRPO & TRiC & Additive & Full VeSTA & Residual \\
\midrule
CollAvg & 0.99 & 6.19 & 7.19 & 7.14 & $-0.05$ \\
CollW & 4.29 & 19.69 & 23.98 & 21.27 & $-2.71$ \\
\bottomrule
\end{tabular}
\caption{Component gains relative to DualVLN and the interaction residual, defined as the full VeSTA gain minus the sum of individual gains. All values are percentage points.}
\label{tab:component_interaction}
\end{table}

For CollAvg, the residual is only $-0.05$ percentage points, so the episode-average effects are nearly additive. For CollW, the $-2.71$-point residual indicates partial overlap or diminishing returns: both mechanisms reduce weighted contact burden, but some contacts removed by one component are no longer available for the other to remove. The evidence therefore supports complementary roles, not a claim of super-additive synergy.

\section{Execution Length and Conservatism}
A trivial way to lower collisions would be to execute fewer actions or stop moving. Table~\ref{tab:execution_length} does not support that universal-shortening explanation. VeSTA has a larger median trajectory length and longer successful trajectories on both benchmarks. Its lower mean length is concentrated in failed episodes and coincides with fewer episodes reaching the 500-step budget.

\begin{table}[H]
\centering
\small
\setlength{\tabcolsep}{5pt}
\begin{tabular}{llrrrrr}
\toprule
Data & Method & Mean & Median & $\geq500$ & Success & Failure \\
\midrule
R2R & DualVLN & 172.88 & 75 & 21.15 & 73.29 & 278.61 \\
R2R & VeSTA & 171.16 & 89 & 15.39 & 100.70 & 252.42 \\
RxR & DualVLN & 254.00 & 152 & 36.49 & 92.93 & 368.44 \\
RxR & VeSTA & 243.04 & 170 & 26.55 & 119.17 & 334.26 \\
\bottomrule
\end{tabular}
\caption{Episode length statistics. Mean, median, successful-episode, and failed-episode columns report steps; $\geq500$ is the percentage of episodes reaching the nominal maximum budget.}
\label{tab:execution_length}
\end{table}

The pattern is consistent with VeSTA completing some successful routes through longer but safer executions while avoiding a subset of very long failure loops. It rules out a universal no-motion explanation, but not every form of conservatism: without matched action-level geometry, the analysis cannot determine whether VeSTA uses smaller local displacements, more rotations, or larger obstacle margins.

\section{Failure Taxonomy and Mechanism Coverage}
Collision aggregates combine several distinct causes. Table~\ref{tab:failure_taxonomy} states which failure modes are directly targeted by the proposed mechanisms and which remain outside their demonstrated scope.

\begin{table}[H]
\centering
\small
\begin{tabularx}{\textwidth}{p{0.23\textwidth}XX}
\toprule
Failure mode & Relation to VeSTA & Remaining limitation \\
\midrule
Semantic planning failure & Neither component directly corrects an incorrect high-level interpretation. & Requires stronger language grounding or global planning. \\
Candidate support failure & TR-GRPO is intended to shift generation toward executable local candidates. & Historical probes do not provide a baseline-versus-TR-GRPO coverage comparison. \\
Candidate selection failure & TRiC ranks available candidates using predicted execution risk and progress. & Logged data do not expose full counterfactual utility for every final-policy decision. \\
Local controller failure & Both components may reduce risky commands presented to the controller. & Neither changes low-level tracking dynamics or guarantees command feasibility. \\
Persistent-contact failure & Strict-contact training and risk-aware selection directly discourage repeated contact. & Minor isolated contacts may remain or be introduced on previously safe episodes. \\
Premature stopping & Positive-progress preferences can discourage non-progress candidates. & Goal-stop calibration is not separately evaluated. \\
Dynamic or unmodelled failure & Not directly addressed by the static simulation training protocol. & Requires sensing and adaptation beyond the evaluated setting. \\
\bottomrule
\end{tabularx}
\caption{Conceptual failure taxonomy and the evidenced scope of the two VeSTA mechanisms.}
\label{tab:failure_taxonomy}
\end{table}

This taxonomy narrows the mechanism claim. VeSTA primarily addresses local candidate support, risk-aware selection, and persistent-contact behavior; it is not a general solution to semantic misunderstanding, controller mismatch, or dynamic obstacles.

\section{Interpretation of the Gaussian Surrogate}
The flow generator does not expose a directly usable exact density over executed trajectory vectors for the policy update. The diagonal Gaussian used by TR-GRPO is therefore a local differentiable interface over controller-facing vectors in $\mathbb{R}^{3H}$, not the exact likelihood of the underlying flow model. Its mean and scale provide a tractable reparameterized update around generated trajectories.

This approximation ignores temporal correlations across the horizon, correlations among coordinates, and multimodality among distinct path families. The current evidence supports its practical utility as an optimization surrogate because the complete intervention improves contact metrics. It does not establish density correctness, calibrated uncertainty, or optimality of the diagonal family. Claims about the surrogate should consequently remain implementation-level rather than probabilistic claims about the flow generator itself.

\section{Collision-Tail Analysis}
Mean collision rate can hide whether an improvement removes severe failures or merely shifts small contacts. Table~\ref{tab:collision_tail} reports episode-level quantiles and threshold exceedance rates.

\begin{table}[H]
\centering
\small
\setlength{\tabcolsep}{4.5pt}
\begin{tabular}{llrrrrrrrr}
\toprule
Data & Method & Median & P75 & P90 & P95 & Free & $>10$ & $>30$ & $>50$ \\
\midrule
R2R & DualVLN & 0.00 & 16.24 & 55.93 & 66.69 & 66.61 & 27.24 & 20.72 & 13.38 \\
R2R & VeSTA & 0.00 & 7.23 & 23.11 & 33.33 & 57.59 & 21.48 & 5.98 & 0.27 \\
RxR & DualVLN & 2.31 & 45.31 & 63.27 & 72.65 & 47.51 & 45.24 & 35.87 & 19.98 \\
RxR & VeSTA & 2.97 & 15.84 & 28.95 & 36.93 & 39.87 & 34.26 & 9.21 & 0.25 \\
\bottomrule
\end{tabular}
\caption{Episode-level collision-rate distribution. Quantiles, collision-free rate (Free), and threshold exceedance rates are percentages. The corresponding means are 13.22/6.08 on R2R and 21.79/9.36 on RxR for DualVLN/VeSTA.}
\label{tab:collision_tail}
\end{table}

VeSTA strongly compresses the severe collision tail. The fraction above 50\% collision rate falls from 13.38\% to 0.27\% on R2R and from 19.98\% to 0.25\% on RxR. However, the collision-free fraction also decreases, from 66.61\% to 57.59\% on R2R and from 47.51\% to 39.87\% on RxR. Thus, the dominant effect is not to make every episode contact-free; it is to convert many severe or persistent-contact failures into lower-contact executions. Longer successful trajectories may also expose more episodes to isolated minor contacts.

\section{Baseline-Risk Stratification}
Ordinary quartiles are unstable here because the baseline contains many exact zero-collision ties. We therefore retain all zero-collision episodes as one stratum and split positive-collision episodes into three equal-count groups after stable sorting by baseline CollAvg.

\begin{table}[H]
\centering
\small
\setlength{\tabcolsep}{6pt}
\begin{tabular}{llrrrrr}
\toprule
Data & Baseline stratum & $N$ & Range & Base & VeSTA & $\Delta$SR \\
\midrule
R2R & Zero & 1225 & 0--0 & 0.00 & 3.97 & $-5.22$ \\
R2R & Positive T1 & 205 & 0.23--24.77 & 10.07 & 5.94 & $-1.95$ \\
R2R & Positive T2 & 205 & 24.80--53.49 & 41.23 & 10.11 & $+23.90$ \\
R2R & Positive T3 & 204 & 53.49--87.82 & 67.65 & 14.85 & $+27.94$ \\
\midrule
RxR & Zero & 1743 & 0--0 & 0.00 & 5.23 & $-9.41$ \\
RxR & Positive T1 & 642 & 0.20--31.48 & 13.94 & 9.65 & $-6.23$ \\
RxR & Positive T2 & 642 & 31.54--52.89 & 43.63 & 12.23 & $+17.76$ \\
RxR & Positive T3 & 642 & 53.09--88.02 & 66.98 & 17.42 & $+19.00$ \\
\bottomrule
\end{tabular}
\caption{Outcomes stratified by baseline episode risk. Range, Base, and VeSTA are CollAvg percentages; $\Delta$SR is VeSTA minus DualVLN in percentage points.}
\label{tab:baseline_risk}
\end{table}

The largest collision reductions and SR gains are concentrated in the medium- and high-risk positive strata. In contrast, episodes that are collision-free under DualVLN acquire 3.97\% CollAvg and lose 5.22 SR points on R2R; the corresponding RxR changes are 5.23\% and $-9.41$ points. The lowest positive-risk strata also show small SR decreases. VeSTA should therefore be described as reducing severe and persistent collision burden, with weaker or adverse effects on episodes that are already safe under the baseline, rather than as uniformly improving every episode.

\section{Additional TRiC Ranking Diagnostics}
We further evaluate ranking quality on the same complete available 419-episode rank-0 shard used in Section A6. Spearman and Kendall coefficients compare each score with realized collision rate, while AUROC measures discrimination of failure, any collision, and collision rate of at least 10\%. These are ranking diagnostics, not probability-calibration results.

\begin{table}[H]
\centering
\small
\setlength{\tabcolsep}{7pt}
\begin{tabular}{lrrrrr}
\toprule
Signal & Spearman & Kendall & Failure AUC & Any-coll. AUC & High-coll. AUC \\
\midrule
Collision risk & 0.655 & 0.532 & 0.794 & 0.876 & 0.860 \\
Stuck risk & 0.522 & 0.420 & 0.867 & 0.804 & 0.766 \\
\bottomrule
\end{tabular}
\caption{TRiC score ranking diagnostics on the historical 419-episode rank-0 shard. High collision denotes realized collision rate of at least 10\%.}
\label{tab:tric_ranking}
\end{table}

The collision-risk score is the stronger collision-ranking signal: its bottom decile has 88.10\% SR and no collisions, whereas its top decile has 4.76\% SR, 33.37\% CollAvg, and 100\% incidence of both any and high collision. The stuck-risk score better discriminates overall failure (AUROC 0.867): its bottom decile has 97.62\% SR and 2.53\% CollAvg, while its top decile has 0\% SR and 30.70\% CollAvg. Because the shard is historical and not assumed to be a random sample of the final evaluation set, these results support qualitative ordering but not calibrated risk probabilities or final-checkpoint generalization.

\section{Scope of the Additional Evidence}
The appendix reports only completed analyses. It does not isolate the group-relative coefficient from the accompanying PPO and strict-contact reward terms, evaluate group-size or penalty sensitivity, or compute final-checkpoint counterfactual candidate utility. Accordingly, the TR-GRPO results are attributed to the complete training intervention, the $G=2$ configuration is interpreted as pairwise group-relative optimization, and the historical candidate and TRiC analyses are treated as descriptive support and ranking diagnostics. The stratified analyses additionally show that improvements are concentrated in initially high-risk episodes and are not uniform. These boundaries prevent unperformed diagnostic experiments from being presented as evidence.

\section{Reproducibility and Claim Boundary}
The numerical simulation analyses in this appendix are generated by \texttt{supplementary\_statistics.py} from existing episode- and decision-level logs. The script fixes the bootstrap seed, recomputes weighted collision rates inside each resample, and implements the tail, stratification, and ranking calculations without external statistical dependencies. The physical-study intervals are computed from the aggregate trial counts reported in the main paper, which are explicitly encoded in the script because disaggregated robot trial records are unavailable. No new checkpoint, citation, robot outcome, or simulator result is introduced by this appendix.

The evidence supports the following scoped interpretation. VeSTA, instantiated on DualVLN, produces large and statistically stable reductions in severe contact burden under the evaluated strict-contact conditions. Small aggregate SR and SPL changes are not uniformly separated from zero, and episode-level benefits are concentrated in medium- and high-risk baseline strata. Historical candidate probes suggest that selection is usually effective when executable support exists, while TRiC outputs rank realized collision and failure risk; neither result establishes calibrated safety probabilities or final-checkpoint counterfactual optimality. The physical study supports transfer directionally, while its sample size remains insufficient for a precise effect estimate. Broader architectural transfer, repeated full training seeds, and expanded robot trials are outside the intended lightweight scope of this appendix.

\end{document}